\documentclass{article}

\PassOptionsToPackage{numbers, sort&compress}{natbib}

\usepackage[final]{nips_2018}

\usepackage[utf8]{inputenc} 
\usepackage[T1]{fontenc}    
\usepackage{hyperref}       
\usepackage{url}            
\usepackage{booktabs}       
\usepackage{amsfonts}       
\usepackage{nicefrac}       
\usepackage{microtype}      
\usepackage{graphicx}
\usepackage{subcaption}

\usepackage{authblk}

\author[1]{Sina Rashidian   }
\author[2]{Janos Hajagos    }
\author[2]{Richard Moffitt  }
\author[1,2]{Fusheng Wang     }
\author[1]{Xinyu Dong       }
\author[2]{Kayley Abell-Hart        }
\author[2]{Kimberly Noel       }
\author[2]{Rajarsi Gupta       }
\author[2]{Mathew Tharakan       }
\author[2]{Veena Lingam       }
\author[2]{Joel Saltz       }
\author[2]{Mary Saltz       }

\affil[1]{Department of Computer Science, Stony Brook University}
\affil[1]{[firstname].[lastname]@stonybrook.edu}
\affil[2]{Department of Biomedical Informatics, Stony Brook Medicine}
\affil[2]{[firstname].[lastname]@stonybrookmedicine.edu}
\affil[ ]{Stony Brook, NY 11794}

\title{Disease phenotyping using deep learning: A diabetes case study}

\begin{document}

\maketitle

\begin{abstract}
Accurate characterization of populations, as captured by International Classification of Diseases (ICD) codes, is key to healthcare decision-making.  Despite being assigned by professionally trained and certified coders, these codes are frequently imprecise.  We present a methodology that uses deep learning to model coder decision-making by assigning ICD codes to inpatient encounters. This creates clinical phenotypes derived from discrete data elements in the medical record. Our approach assigns codes based on demographics, lab results, and medications, as well as codes from previous encounters. Using diabetes as a test case, the model was able to predict existing codes with high accuracy. We then employed a panel of practicing physicians who, in a blinded manner, established an independent ground truth which we used to assess the accuracy of our model. In cases where the model strongly disagreed with the existing codes, the physician review agreed overwhelmingly with the model. Furthermore, the model-predicted probability of diabetes tracked closely to the physician-established likelihood of diabetes. Our data suggests that 9.07\% of cases had missing or incorrect ICD codes for diabetes.
\end{abstract}

\section{Introduction} \label{Introduction}


Accurate identification, documentation, and coding of disease is important to health care, relating directly to patient care, revenue, and performance evaluation. ICD (International Classification of Diseases) codes are used to classify mortality, define cohorts, evaluate health care policy, and drive health care finance. Despite their importance, there is considerable inaccuracy and variability in these assigned codes \citep{o2005measuring, geruso2015upcoding}.
Recently, there has been high interest in the use of deep learning methods to overcome obstacles to working with Electronic Health Records (EHR) data; specifically, focused on prediction of patient diagnosis, readmission and mortality \citep{lipton2015learning, esteban2016predicting, choi2016doctor, rajkomar2018scalable, avati2017improving, nguyen2017finding, miotto2016deep}. More applications of deep learning in EHR are discussed in survey works \citep{shickel2018deep, miotto2017deep} thoroughly.

In this paper, we tackle the problem of disease phenotyping – generating EHR based diagnosis codes or flagging existing codes which may need revision. For this purpose, first we train a deep learning model to mimic coder behavior for assigning diagnosis codes based on discrete data available for each patient’s visit. After optimizing this model, we study cases where the model prediction was different from the coders’ annotation. We then analyze these discordant cases by sampling from different intervals of the model's output probability, followed by reviewing these cases with a group of experts. In this work, we focus on diabetes for analyzing discordant cases in detail, but the model and training methodology do not contain any disease-specific choices.

\section{Data}

Data for this study were extracted from the Cerner HealthFacts database, a large multi-institutional de-identified database derived from EHRs and administrative systems. There are 599 facilities in this database, consisting of both inpatient and outpatient encounters. We selected cohorts of inpatient encounters: (1) from large volume acute-care facilities, (2) from facilities which report laboratory tests and diagnosis codes, and (3) of patients 50 years and older (who are majority of diabetes cases). The main dataset for this manuscript is from facility 143, which was chosen randomly from the top 10 volume facilities in this database for the time period 8/24/2006 to 12/31/2013. 

After mapping all extracted data to the OHDSI Common Data Model (version 5.3), we combined all available data of each encounter, including demographic information, laboratory results, medications and observations for creating our feature matrix. All categorical features were converted to vector format using a 1-hot encoding scheme. Our target phenotype was a combination of all diabetes diagnosis codes (defined by CCS codes 49 \& 50), meaning if any ICD code related to diabetes is reported, the patient is diabetic and is a positive case in our feature matrix. Since the number and frequency of each laboratory result varies between different encounters, we aggregated all of the values from each test in statistics elaborated in Table \ref{table_aggregates}. 

\begin{table}
  \caption{Aggregated statistics for each specific test in each encounter }
  \label{table_aggregates}
  \centering
  \begin{tabular}{ll}
    \toprule
    Statistics              &   Description \\
    \midrule
    Count                   & Number of times test was ordered \\
    Min                     & Minimum value between all values\\
    Max                     & Maximum value between all values   \\
    Median                  & Median of all values   \\
    Values' concepts        & Number of times values fall in high, normal or low\\
    \enspace              & ranges defined by the facility\\
    Delta                   & Last value minus first value\\

    \bottomrule
  \end{tabular}
\end{table}

\section{Methodology} \label{Methodology}
Given that many patient tests are unresulted, we reduce the sparsity of our data  by removing irrelevant features. Since many tests are unrelated to our target disease, we only keep features that are common in at least 5\% of positive cases in the training set. By doing so, we reduced the number of features from 13,139 to 966. 
Diagnosis codes are assigned based on a specific set of guidelines and rules. By investigating thousands of training data points, we expect a deep learning model to discover such patterns. Deep Learning models are ideal for these tasks, as they do not require a separate feature selection step and they perform well with a relatively large feature space. In this work, we decided to use a multilayer perceptron network. More than 2/3 of patients had no previous encounter information; therefore, a time series model like a recurrent neural network was not suitable. 
We chose to select a modeling framework that ran well on the majority of patients where no prior information was available, reflecting a more real-world application. Furthermore, laboratory results do not provide clinical value after a long time period, and the salient information about patient history is better captured in disease codes from previous encounters. As a result, we aggregated previous encounter information into one binary vector of diagnosis codes, representing whether a patient had that disease code in any previous encounter. We applied this strategy for all individual diagnosis codes, as well as for the aggregated code for the target disease - diabetes.

\subsection{Neural Network Architecture}
We implemented a multilayer perceptron Deep Neural Network (DNN) for this study. The input layer has 966 features and the output is a single neuron consisting of sigmoid activation function, assigning probability to each input vector. We performed extensive hyperparameter tuning over a variety of activation functions (tanh, ReLU and SeLU), optimizers (Adam, SGD), loss functions (mean squared error, mean absolute error, binary cross-entropy and categorical hinge), and numbers of hidden layers (ranging 2 to 15). A network with 10 hidden layers consisting of tanh activation function using Adam optimizer and mean squared error as loss function achieved the best result. To avoid overfitting, L1 regularization and dropout were employed. 

To train the deep learning model, we used the Python programming language (2.7), Keras framework with underlying Tensorflow, and scikit learn library. The training was performed on a single computer with a  NVIDIA Tesla V100 (16GB RAM) \citep{chollet2015keras, pedregosa2011scikit, abadi2016tensorflow}.

\section{Results} \label{Results}

\begin{table}
  \caption{Comparison with machine learning models}
  \label{table_compare2ML}
  \centering
  \begin{tabular}{lcccccc}
    \toprule
    Algorithm               &   Precision   & Recall & F1-Score &  AUCROC & AP \\
    \midrule
    Deep Neural Network     &   0.79 &	0.82 &	0.80 &	0.92 &	0.81  \\

    Logistic Regression     &   0.75 &	0.79 &	0.77 &	0.90 &	0.78  \\
    Random Forest           &   0.80 &	0.66 &	0.72 &	0.90 &	0.77  \\
    LinearSVC               &   0.74 &	0.75 &	0.75 &	0.82 &	0.63  \\
    \bottomrule
  \end{tabular}
\end{table}

\begin{figure}
  \begin{subfigure}[b]{0.50\textwidth}
      \includegraphics[width=\textwidth]{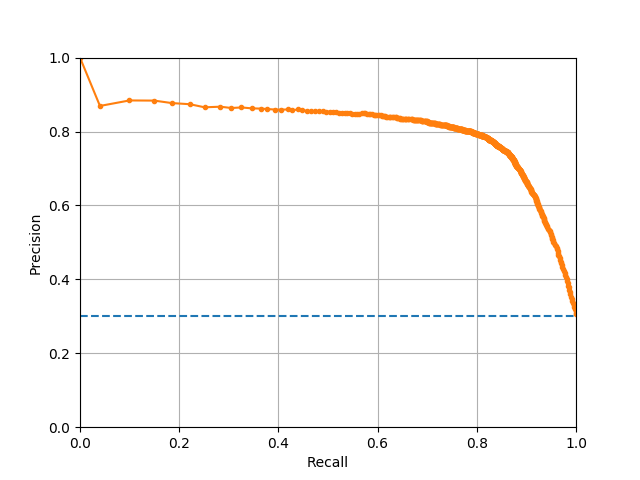}
      \caption{Precision recall curve}\label{fig_auprc}
  \end{subfigure}
    ~
  \begin{subfigure}[b]{0.50\textwidth}
      \includegraphics[width=\textwidth]{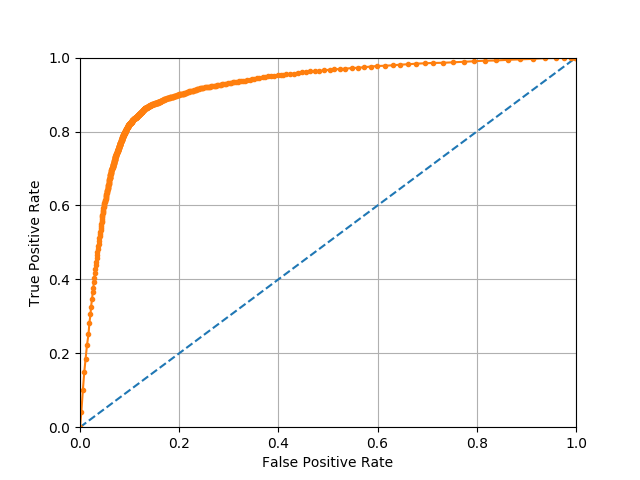}
      \caption{Receiver operating characteristic curve}\label{fig_auroc}
  \end{subfigure}
  \caption{DNN model performance analysis on the test set}\label{fig_curves}
\end{figure}

\subsection{Deep learning method performance}
20\% of data was untouched for test set and the rest was divided into training and validation sets. All results are based on test set.
We compared the trained DNN to three popular machine learning methods. As shown in Table \ref{table_compare2ML}, the deep learning method surpassed the other machine learning algorithms in all aggregated metrics: F1-Score, AUC-ROC, and Average Precision. Since our dataset is imbalanced (30\% prevalence), we prioritized the F1-score which balances precision and recall. 

In Figure \ref{fig_curves}, we demonstrate how the model performs across different levels of model confidence. For comparison, dashed lines show how a random classifier would perform. Notably, our model maintained a precision above 80\% until recall reached nearly 80\%.

\begin{figure}
  \begin{subfigure}[b]{0.50\textwidth}
      \includegraphics[width=\textwidth]{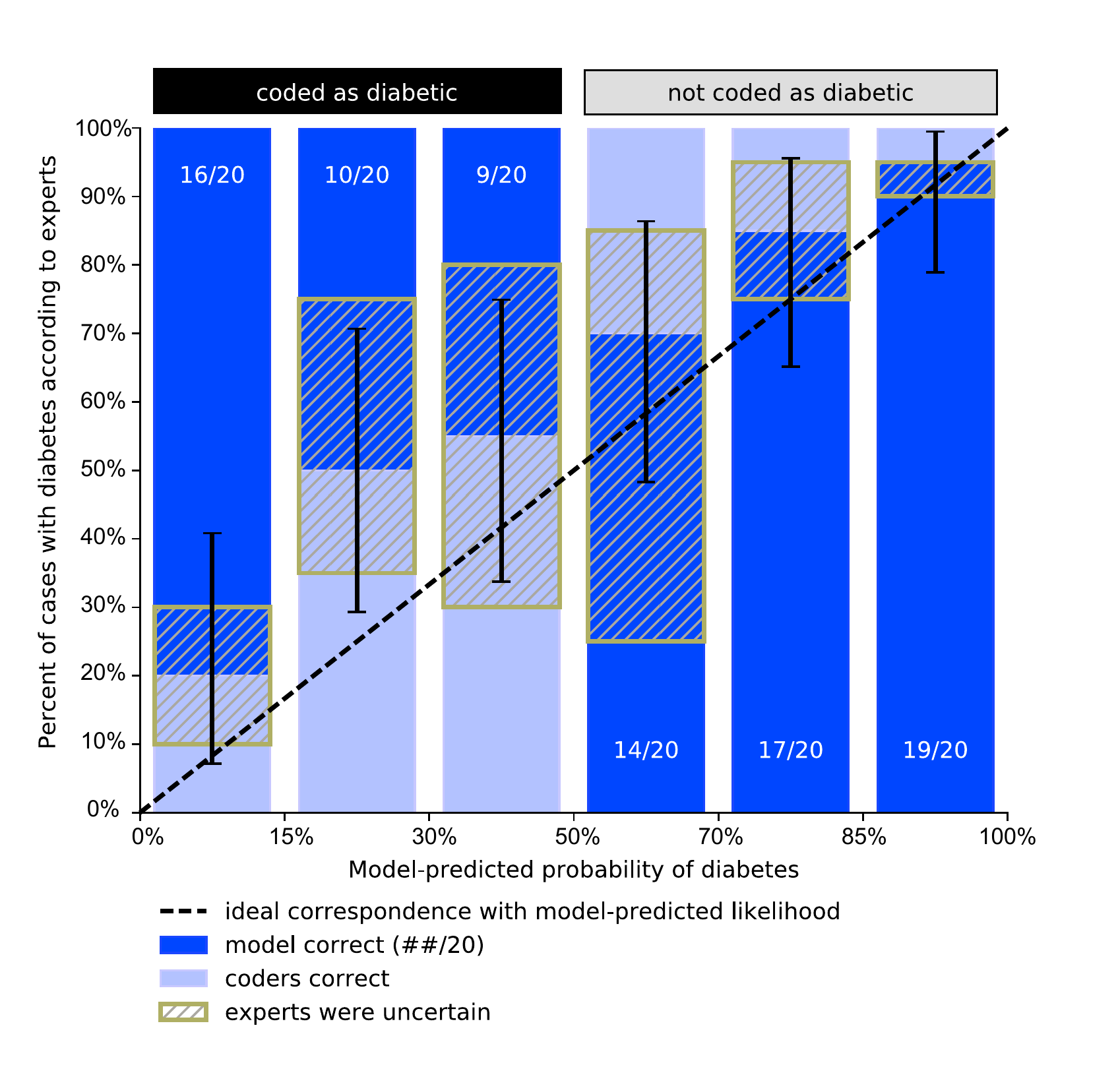}
    \caption{}
      \label{fig_experts}
  \end{subfigure}
  ~
  \begin{subfigure}[b]{0.50\textwidth}
      \includegraphics[width=\textwidth]{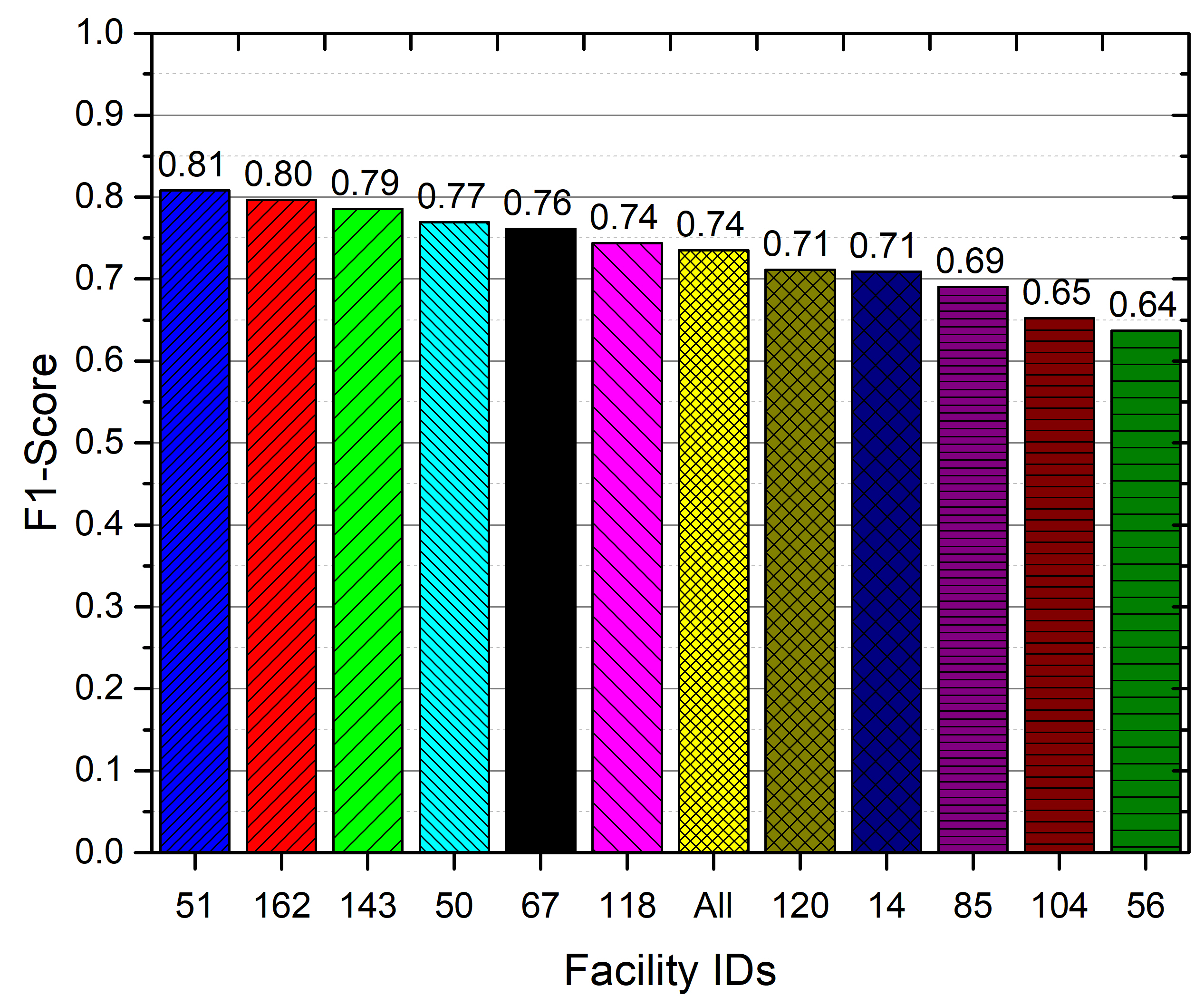}
    \caption{}
    \label{fig_multi_facility}
  \end{subfigure}
 
 \caption{a. Summary of expert review in cases where the coders and model disagreed. Dark blue color reflects cases in which the experts agreed with the model prediction, and vertical error bars show the 95\% confidence interval for the expert-provided rate of diabetes in each set of cases. Grey hashed areas demonstrate cases where the experts stated low-confidence in their assessments. b. F1-score across 11 different facilities for a model trained with data from all 11 facilities.}
\end{figure}

\subsection{Analyzing model-coder discordant cases}
We expected diabetes prediction to be relatively straightforward, given that many lab values and medications are specific to diabetes. However, when validated against the coders’ diagnoses,  the model performance was lower than anticipated. Since inconsistency in disease codes was a motivation for this study, we asked a group of three physicians to review cases in which the model and coders disagreed. In addition to determining whether each case should or should not be coded for diabetes, the physicians also noted their own confidence (high or low) in this assessment. This review was conducted in double-blind fashion, as both the physicians as well as the scientist managing the review process were unaware of the model and coder labels.
As the output of the last layer in deep learning is a sigmoid function, it is analogous to the model’s assessed probability of being diabetic for each input vector ($p$). Discordant cases were grouped based on their probability ($p$) into one of three bins: High confidence ($p<0.15$ or $0.85<p$) medium confidence ( $0.15<=p<0.3$ or $0.70<p<=0.85$) and low confidence ($0.3<=p<=0.7$). For each confidence interval, we sampled 40 cases (20 each from cases in which diabetes was coded but $p<0.5$, or for which diabetes was not coded but $p>0.5$) for review. 
Expert review, shown in Figure \ref{fig_experts}, suggested that the model-generated probability ($p$) of diabetes was similar to the actual prevalence of diabetes in each group of cases, meaning that the coders were wrong at a frequency predicted by the model’s confidence. The coders were especially incorrect in cases when the model disagreed with high confidence (35/40). Notably, in 16 out of 20 cases where coders had documented the presence of diabetes, physicians agreed with the model there was no evidence of disease. Furthermore, among cases that the model flagged with low confidence, there was also a decrease in the physicians’ confidence.

\subsection{Multi-Facility expanding}
In order to improve the generalizability of our procedure to other facilities, we repeated training the DNN with combined data from 11 facilities. Ideally, the model should learn cross-facility differences automatically. The multi-facility data consisted of balanced sampling from 11 facilities shuffled randomly without any facility ID during training. Aside from this new training data, the other training procedures were the same as described above. The performance of this model on independent test data from each of these 11 facilities is shown in Figure \ref{fig_multi_facility}. The F1-score for facility 143 when trained data included 10 other facilities is 0.79 which is very close to what we achieved when we trained and tested on facility 143 alone (0.80). This suggests that our deep learning strategy has the capability to account for facility-specific batch effects. Furthermore, for facilities other than 143, results were still reasonably good, especially when considering the rate at which coder-generated training labels were shown to be inaccurate in the data from facility 143. This experiment shows this methodology to be scalable and generalizable among multiple facilities to create one single model for coding diabetes.

\subsection{Effect on population prevalence}
The expert re-analysis of discordant cases gave us the capability to make a conservative estimate of global miscoding rates for diabetes (as even concordant cases could be incorrect). Our test set consisted of 16,797 encounters, 2,082 of which the model disagreed with the coders. From these 2,082 disagreements, the model had high confidence in 748 cases, medium confidence in 787 cases, and low confidence in 547 cases. Based on our experts experiment and projecting rates onto all of the discordant tests suggests that about 1,523 cases are incorrectly coded, i.e. 73\% of the discordant cases, and 9.07\% of the total population. From these 9.07\%, there are 5.68\% missing diabetes codes, and 3.39\% where a diabetes diagnosis code is wrongly assigned to the patient and should be removed.

\section{Conclusion}
A trained deep learning model was successful at reproducing coder-documented presence or absence of diabetes. Our model is notable, in that it was built with minimal design choices, did not require medical experts to define feature and is expected to generalize to other diseases. As we built our model on data mapped to a widely utilized clinical OHDSI CDM (Common Data Model) issues of data interoperability across institutions are minimized.

The manual review of results confirmed that the model’s performance was concordant with that of the medical experts. Data suggests that up to 9.07\% of encounters are incorrectly coded for diabetes. Improved accuracy of diabetic patients, should assist health care systems improve patient registries and may help them identify patients requiring intensive follow-up care.

The model performed well when challenged with data from multiple institutions, suggesting that with a sufficiently large training set, we might be able to deploy the model to institutions not included in the training set.

\bibliographystyle{ACM-Reference-Format}

\bibliography{sigproc}
\end{document}